\documentclass[twoside,12pt]{article}
\usepackage{array,times,amsmath,amsfonts,amsthm,amssymb,eucal}
\usepackage{graphicx,ifthen}
\usepackage{wrapfig, framed, caption}
\usepackage{latexsym}
\usepackage{color}
\usepackage{mathrsfs}
\usepackage{bm}
\usepackage{hyperref}
\usepackage{subfigure}
\usepackage[authoryear]{natbib}
 \usepackage[normalem]{ulem}
 \useunder{\uline}{\ul}{}
 \usepackage{adjustbox,lipsum}
 \usepackage{booktabs}
 \usepackage{graphicx}

\usepackage{algorithm}
\usepackage{algorithmic}
\usepackage{caption}
\usepackage{tikz}
\usetikzlibrary{calc}
\tikzstyle{every node}=[font=\large]
\tikzstyle{every path}=[line width=2pt]

\makeatletter
\def\BState{\State\hskip-\ALG@thistlm}
\makeatother

\begin{document}
\DeclareGraphicsExtensions{.gif,.pdf,.png,.jpg,.tiff}

\def\({\left(}
\def\){\right)}
\def\[{\left[}
\def\]{\right]}
\def\a{\alpha}\def\b{\beta}\def\d{\delta}\def\D{\Delta}
\def\e{\varepsilon}\def\g{\gamma}\def\G{\Gamma}\def\k{\kappa}
\def\l{\lambda}\def\L{\Lambda}\def\m{\mu}\def\p{\phi}\def\P{\Phi}
\def\r{\rho}\def\s{\sigma}\def\t{\theta}\def\T{\Theta}
\def\ta{\tau}\def\z{\zeta}
\def\RSF{\mathscr}
\def\aa{{\RSF A}}\def\bb{{\RSF B}}\def\cc{{\RSF C}}
\def\dd{{\RSF D}} \def\ee{{\RSF E}}\def\ff{{\RSF F}}\def\gg{{\RSF G}}
\def\hh{{\RSF H}} \def\ii{{\RSF I}}\def\jj{{\RSF J}}\def\kk{{\RSF K}}
\def\ll{{\RSF L}} \def\mm{{\RSF M}}\def\nn{{\RSF N}}\def\oo{{\RSF O}}
\def\pp{{\RSF P}} \def\qq{{\RSF Q}}\def\rr{{\RSF R}}\def\ss{{\RSF S}}
\def\tt{{\RSF T}} \def\uu{{\RSF U}}\def\vv{{\RSF V}}\def\ww{{\RSF W}}
\def\xx{{\RSF X}} \def\yy{{\RSF Y}}\def\zz{{\RSF Z}}
\font\Caps=cmcsc10 \font\BigCaps=cmcsc12 scaled \magstep 1
\font\BigSlant=cmsl10    scaled \magstep 1
\font\proclaimfont=cmbx9 scaled \magstep 1
\def\BigHeading{\bfseries\Large}\def\MediumHeading{\bfseries\large}
\def\smc{\Caps}
\def\lbk{\linebreak}
\newdimen\bigindent
\newdimen\smallindent
\bigindent=30pt \smallindent=5pt
\def\quoteindent{\advance\leftskip by\bigindent\advance\rightskip
                 by\bigindent}
\newskip\proclaimskipamount
\proclaimskipamount=12pt  plus1pt minus1pt
\def\proclaimskip{%
  \par\ifdim\lastskip<\proclaimskipamount
  \removelastskip\vskip\proclaimskipamount\fi}
\let\demoskip=\proclaimskip
\def\Demo#1{\par\ifdim\lastskip<\proclaimskipamount
            \removelastskip\proclaimskip\fi
            \noindent\sl#1. \hskip\smallindent\rm}
\def\EndDemo{\par\demoskip}
\def\DemoSection#1{\par\ifdim\lastskip<\proclaimskipamount
             \removelastskip\proclaimskip\fi
             #1\hskip\smallindent\rm}
\def\Section#1{\stepcounter{section}
    \DemoSection{{\bfseries\large\thesection.\hskip\smallindent#1}}}
\def\Subsection#1{\stepcounter{subsection}
    \DemoSection{\bfseries\normalsize\thesubsection.\hskip\smallindent#1}}
\def\Quote{\begin{quotation}\normalfont\small}
\def\EndQuote{\end{quotation}\rm}

\newtheorem{theorem}{Theorem}
\newtheorem{lemma}{Lemma}
\newtheorem{corollary}{Corollary}
\newtheorem{remark}{Remark}
\newtheorem{example}{Example}
\newtheorem{definition}{Definition}
\def\Theorem{\begin{theorem}\sl}
\def\EndTheorem{\end{theorem}}
\def\Lemma{\begin{lemma}\sl}
\def\EndLemma{\end{lemma}}
\def\Corollary{\begin{corollary}\sl}
\def\EndCorollary{\end{corollary}}
\def\Remark{\begin{remark}\rm}
\def\EndRemark{\end{remark}}
\def\Example#1{\begin{example}(#1).\hskip\smallindent\rm}
\def\EndExample{\end{example}}
\def\Definition{\vskip10pt\begin{definition}}
\def\EndDefinition{\end{definition}}

\def\bct{\begin{center}}
\def\ect{\end{center}}
\def\Array{\begin{eqnarray*}}
\def\EndArray{\end{eqnarray*}}
\def\Enumerate{\begin{enumerate}}
\def\EndEnumerate{\end{enumerate}}
\def\Itemize{\begin{itemize}}
\def\EndItemize{\end{itemize}}
\def\Eq{\begin{equation}}
\def\EndEq{\end{equation}}
\def\EqArray{\begin{eqnarray}}
\def\EndEqArray{\end{eqnarray}}
\def\mref#1{(\ref{#1})}
\def\qt#1{\qquad\text{#1}}
\def\Tabular{\begin{tabular}}
\def\EndTabular{\end{tabular}}
\def\FlushLeft{\begin{flushleft}}
\def\EndFlushLeft{\end{flushleft}}
\def\FlushRight{\begin{flushright}}
\def\EndFlushRight{\end{flushright}}
\newcommand{\red}{\textcolor{red}}
\newcommand{\blue}{\textcolor{blue}}
\newcommand{\green}{\textcolor{green}}
\def\pmb#1{\setbox0=\hbox{#1}%
 \kern-0.010em\copy0\kern-\wd0
 \kern0.035em\copy0\kern-\wd0
 \kern0.010em\raise.0233em\box0}

\def\bhat{\hat{\beta}}
\def\bbag{{\hat{\beta}_{\text{inbag}}}}
\def\err{\text{Err}}
\def\oob{\text{oob}}
\def\marg{{\!\text{marg}}}
\def\step{\text{step}}
\def\shat{\hat{\sigma}}
\def\vimp{\Delta}

\newcolumntype{L}{>$l<$}
\newcolumntype{L}{>$c<$}


\def\Report{A Machine Learning Alternative to P-values}
\def\Author{Min Lu and Hemant Ishwaran}
\pagestyle{myheadings}\markboth{\Author}{\Report}
\thispagestyle{empty}

\bct{\BigHeading A Machine Learning Alternative to P-values\vskip10pt} 
Min Lu and Hemant Ishwaran\lbk 
Division of Biostatistics,
University of Miami\lbk
\rm\today 
\ect

\noindent
This paper presents an alternative approach to p-values in
regression settings.  This approach, whose origins can be traced to
machine learning, is based on the leave-one-out bootstrap for
prediction error.  In machine learning this is called the out-of-bag
(OOB) error.  To obtain the OOB error for a model, one draws a
bootstrap sample and fits the model to the in-sample data.  The
out-of-sample prediction error for the model is obtained by
calculating the prediction error for the model using the out-of-sample
data.  Repeating and averaging yields the OOB error, which represents
a robust cross-validated estimate of the accuracy of the underlying
model.  By a simple modification to the bootstrap data involving
``noising up'' a variable, the OOB method yields a variable importance
(VIMP) index, which directly measures how much a specific variable
contributes to the prediction precision of a model.  VIMP provides a
scientifically interpretable measure of the effect size of a
variable, we call the {\it predictive effect size}, that holds whether
the researcher's model is correct or not, unlike the p-value whose
calculation is based on the assumed correctness of the model.  We also
discuss a marginal VIMP index, also easily calculated, which
measures the marginal effect of a variable, or what we call the {\it
  discovery effect}.  The OOB procedure can be applied to both
parametric and nonparametric regression models and requires only that
the researcher can repeatedly fit their model to bootstrap and
modified bootstrap data.  We illustrate this approach on a survival
data set involving patients with systolic heart failure and to a
simulated survival data set where the model is incorrectly specified
to illustrate its robustness to model misspecification.


\section{Introduction}

The issue of p-values has taken center stage in the media with many
scientists expressing grave concerns about their validity.  ``P
values, the 'gold standard' of statistical validity, are not as
reliable as many scientists assume'', is the leading assertion of the
highly accessed {\it Nature} article, ``Scientific method: Statistical
errors''~\citep{nature}.  Even more extreme is the recent action of
the journal of Basic and Applied Social Psychology (BASP), which
announced it would no longer publish papers containing p-values.  In
explaining their decision for this policy~\citep{ban-pvalue}, the
editors stated that hypothesis significance testing procedures are
invalid, and that p-values have become a crutch for scientists dealing
with weak data.  These, and other highly visible discussions, so
alarmed the American Statistical Association (ASA), that it recently
issued a formal statement on p-values~\citep{asa}, the first time in
its history it had ever issued a formal statement on matters of
statistical practice.

A big part of the problem is that researchers want the p-value to be
something that it was never designed for.  Researchers want to make
context specific assertions about their findings; they especially want
a statistic that allows them to assert statements regarding scientific
effect.  Because the p-value cannot do this, and because the
terminology is confusing and stifling, this leads to misuse and
confusion.  Another problem is verifying correctness of the model
under which the p-value is calculated.  If model assumptions do not
hold, the p-value itself becomes statistically invalid.  This is not
an esoteric point.  Commonly used models such as linear regression,
logistic regression, and Cox proportional hazards can involve strong
assumptions.  Common practices such as fitting main effect models
without interactions, assuming linearity of variables, and invoking
distributional assumptions regarding the data, such as normality, can
easily fail to hold.  Moreover, the functional relationship between
attributes and outcome implicit in some of these models, such as
proportionality of hazards, may also fail to hold.  Researchers rarely
test for model correctness, and even when they do, they invariably do
so by considering goodness of fit.  But goodnesss of fit measures are
notoriously unreliable for assessing the validity of a
model~\citep{two-cultures}.  All of this implies that a researchers'
findings, which hinges so much on the p-value being correct, could
be suspect without their even knowing it.  This fragility of the p-value
is further compounded by other conditions typically outside of the
control of the researcher, such as the sample size, which has 
enormous effect on its efficacy.

In this paper we focus on the use of p-values in the context of
regression models.  All
widely used statistical software provide p-value information when
fitting regression models; typically p-values are given for the
regression coefficients.  These are provided in an ANOVA table with
each row of the table displays the regression coefficient estimate,
$\bhat$, for a specific coefficient, $\b$, an estimate of its standard
error, $\shat_\b$, and then finally the p-value of the coefficient,
obtained typically by comparing a $Z$-statistic to a normal
distribution:
$$
Z_{\text{observed}} = \frac{\bhat}{\shat_\b},\hskip10pt
\text{p-value}=P\{Z\ge |Z_{\text{observed}}|\}.
$$  
The p-value for the regression coefficient represents the statistical
significance of the test of the null hypothesis $H_0\!\!:\b=0$.  In
other words, it provides a means of assessing whether a specific
coefficient, in this case $\b$, is zero. However, there is a subtle
aspect to this where confusion can take place.  When considering
this p-value, it is important to keep in mind that its value is
calculated not only under the null hypothesis of a zero coefficient value,
but also assuming that {\it the model holds}.
Thus, technically speaking, the null hypothesis is not just that the
coefficient is zero, but is a collection
of assorted assumptions, which should probably read something like:
$$
H_0\!: \,
\Bigl\{\b = 0, \hskip3pt
\text{model holds},\hskip3pt
\text{model assumptions hold (e.g.\, interactions not present)}\Bigr\}.
$$
If any of these assumptions fail to hold, then the p-value is
technically invalid.

\subsection{Contributions and outline of the article}

Given these concerns with the p-value, we suggest a different approach
using a quantity we call the variable importance (VIMP) index.  Our
VIMP index is based on variable importance, an idea that originates
from machine learning.  One of its earliest examples can be traced to
Classification and Regression Trees (CART), where variable importance
based on surrogate splitting was used to rank variables~(see Chapter 5
of~\cite{CART}).  The idea was later refined for variable selection in
random forest regression and classification models by using prediction
error~\citep{two-cultures, random.forest}.  Extensions to random
survival forests were considered by~\cite{rsf}.  Our VIMP index uses
the same idea as these latter approaches, but recasts it within the
p-value context.  Like those methods, it uses prediction error to
assess the effect of a variable in a model.  It replaces
the statistical significance of a p-value with the predictive
importance of a variable.  Most importantly, the VIMP index holds
regardless of whether the model is true.  This is because the index is
calculated using test data and is not based on a presupposed model
being true as the p-value does.

In statistics, effect size is a quantitative measure of the strength
of a phenomenon, which includes as examples: Cohen's $d$ (standard
group mean difference); the correlation between two variables; and
relative risk. In regression models, effect size is measured by the
standardized $\bhat$ coefficient.  Since VIMP is also a measure of the
quantitative strength of a variable, we refer to its quantitative
measure as {\it predictive effect size} to prevent readers from
confusing it with the traditional effect size.  With a simple
modification to the VIMP procedure, we estimate another quantity we call
marginal VIMP and refer to its quantitative measure as the {\it
  discovery effect size}.  This refers to the discovery contribution
of a variable, which will be explained in Section 4.  An important
aspect of both our procedures is that they can be carried out using the
same models the researcher is interested in studying.  Implementing
them only requires the ability to resample the data, apply some
modifications to the data, and calculate prediction error.  Thus they
can easily be incorporated with most existing statistical software
procedures.

Section 2 outlines the VIMP index and provides a formal algorithmic
formulation (see Algorithm~\autoref{A:OOB.VIMP}).  The VIMP index is
based on out-of-bag (OOB) estimation, which relies on bootstrap
sampling. These concepts are also discussed in Section 2.  Section 3
illustrates the use of the VIMP index to a survival data set involving
patients with systolic heart failure with cardiopulmonary stress
testing.  We show how to use this value to rank risk factors and
assess their predictive effect sizes.  In Section 4 we discuss the
extension to marginal VIMP (Algorithm~\autoref{A:OOB.marginal.VIMP})
and show how this can be used to estimate discovery effect sizes in
the systolic heart failure example.  Section 5 studies how sample size
($n$) effects VIMP, comparing this to p-values to show robustness of
VIMP to $n$, then in Section 6 we use a synthetically constructed data
set where the model is incorrectly specified to illustrate the
robustness of VIMP in misspecified settings.  We conclude the paper
with a discussion in Section 7.

\section{OOB prediction error and VIMP}

OOB estimation is a bootstrap technique for estimating the prediction
error of a model.  While the phrase ``out-of-bag'' might be unfamiliar to
readers, the technique has been known for quite some time in the
literature, appearing under various names and seemingly different
guises.  In the statistical literature, the OOB estimator is refered
to as the {\it leave-one-out bootstrap} due to its connection to 
leave-one-out cross-validation~\citep{632}.  See also the earlier
paper by~\cite{efron:error-rate} where a similar idea is discussed.
It is also used in machine learning where it is refered to
as OOB estimation~\citep{oob} due to its connections to the machine
learning method, bagging~\citep{bagging}.

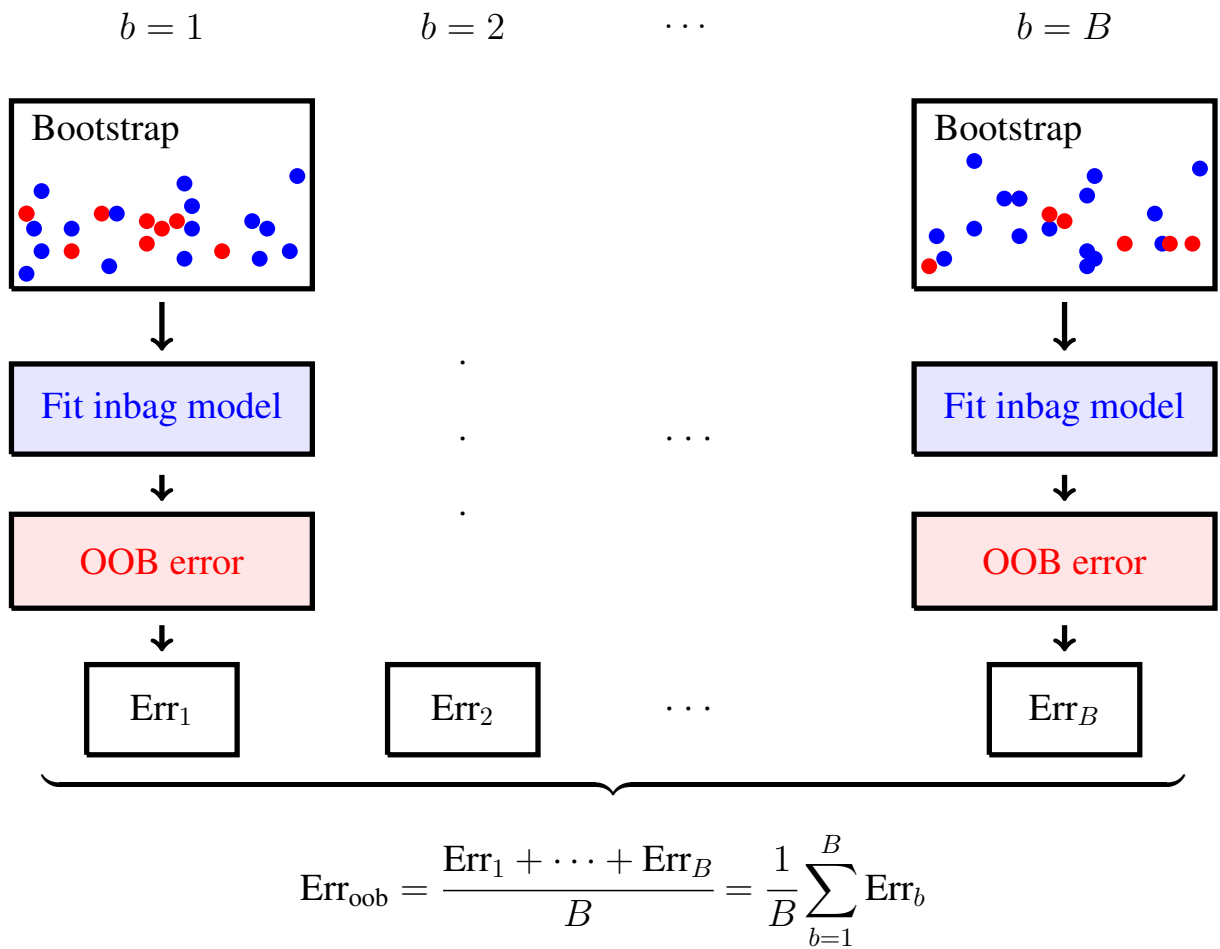
\begin{figure}[phtb]
\hskip-22pt
  \centering
  \begin{tikzpicture}[node distance=0.2cm]
      \draw[fill=white] 
      (0,0)  -- (4,0) -- (4,2.5) -- (0,2.5)--(0,0);  
      \fill[red] (0.2,1)  circle[radius=3pt];
      \fill[blue] (0.4,1.3)  circle[radius=3pt];
      \fill[red] (0.8,0.5)  circle[radius=3pt];
      \fill[blue] (1.3,0.3)  circle[radius=3pt];
      \fill[red] (1.2,1)  circle[radius=3pt];
      \fill[blue] (1.4,1)  circle[radius=3pt];
      \fill[red] (1.8,0.6)  circle[radius=3pt];
      \fill[blue] (2.3,0.4)  circle[radius=3pt];
      \fill[red] (2.2,0.9)  circle[radius=3pt];
      \fill[blue] (2.4,0.8)  circle[radius=3pt];
      \fill[red] (2.8,0.5)  circle[radius=3pt];
      \fill[blue] (3.3,0.4)  circle[radius=3pt];
      \fill[blue] (3.2,0.9)  circle[radius=3pt];
      \fill[blue] (3.4,0.8)  circle[radius=3pt];
      \fill[blue] (3.8,1.5)  circle[radius=3pt];
      \fill[blue] (3.7,0.5)  circle[radius=3pt];
      \fill[blue] (0.2,0.2)  circle[radius=3pt];
      \fill[blue] (0.4,0.5)  circle[radius=3pt];
      \fill[blue] (0.8,0.8)  circle[radius=3pt];
      \fill[blue] (0.3,0.8)  circle[radius=3pt];
      \fill[red] (2,0.8)  circle[radius=3pt];
      \fill[blue] (2.4,1.1)  circle[radius=3pt];
      \fill[red] (1.8,0.9)  circle[radius=3pt];
      \fill[blue] (2.3,1.4)  circle[radius=3pt];
            \draw[fill=white] 
      (12,0)  -- (16,0) -- (16,2.5) -- (12,2.5)--(12,0);  
            \fill[red] (0.2,1)  circle[radius=3pt];
      \fill[red] (13.4,1.2)  circle[radius=3pt];
      \fill[blue] (12.8,0.8)  circle[radius=3pt];
      \fill[blue] (14.3,0.5)  circle[radius=3pt];
      \fill[blue] (13.2,1.2)  circle[radius=3pt];
      \fill[blue] (14.4,1.5)  circle[radius=3pt];
      \fill[blue] (13.8,0.8)  circle[radius=3pt];
      \fill[blue] (15.3,0.6)  circle[radius=3pt];
      \fill[blue] (14.4,0.4)  circle[radius=3pt];
      \fill[blue] (13.4,0.7)  circle[radius=3pt];
      \fill[red] (14.8,0.6)  circle[radius=3pt];
      \fill[blue] (14.3,0.3)  circle[radius=3pt];
      \fill[blue] (15.2,1)  circle[radius=3pt];
      \fill[red] (15.4,0.6)  circle[radius=3pt];
      \fill[blue] (15.8,1.6)  circle[radius=3pt];
      \fill[red] (15.7,0.6)  circle[radius=3pt];
      \fill[red] (12.2,0.3)  circle[radius=3pt];
      \fill[blue] (12.4,0.4)  circle[radius=3pt];
      \fill[blue] (12.8,1.7)  circle[radius=3pt];
      \fill[blue] (12.3,0.7)  circle[radius=3pt];
      \fill[red] (14,0.9)  circle[radius=3pt];
      \fill[blue] (13.4,1.2)  circle[radius=3pt];
      \fill[red] (13.8,0.99)  circle[radius=3pt];
      \fill[blue] (14.3,1.24)  circle[radius=3pt];
      \node (t) at (2,3.5) {$b=1$};
      \node (t) at (6,3.5) {$b=2$};
      \node (t) at (9,3.5) {$\dots$};
      \node (t) at (14,3.5) {$b=B$};
      \node (t) at (6,-3) {$\cdot$};
      \node (t) at (6,-1) {$\cdot$};
      \node (t) at (6,-2) {$\cdot$};
      \node (t) at (9,-2) {$\cdot\cdot\cdot$};
      \node (t) at (9,-5.6) {$\cdot\cdot\cdot$};
      \node (t0) at (1.25,2.1) {Bootstrap};
      \node (t0) at (13.25,2.1) {Bootstrap};
      \node (t1) at (2,0) {};
      \node (t2) at (2,-1) {};
      \node (t11) at (14,0) {};
      \node (t12) at (14,-1) {};
      \draw[fill=blue!10] 
      (0,-2.2)  -- (4,-2.2) -- (4,-1) -- (0,-1)--(0,-2.2); 
      \node[blue] (s) at (2,-1.6) {Fit inbag model};
      \node (s0) at (2,-2.3){};
      \node (s1) at (2,-3) {};
       \draw[fill=blue!10] 
      (12,-2.2)  -- (16,-2.2) -- (16,-1) -- (12,-1)--(12,-2.2); 
      \node[blue] (s) at (14,-1.6) {Fit inbag model};
      \node (s10) at (14,-2.3){};
      \node (s11) at (14,-3) {};
      \draw[fill=red!10] 
      (0,-4.25)  -- (4,-4.25) -- (4,-3) -- (0,-3)--(0,-4.25); 
      \node[red] (r) at (2,-3.625) {OOB error};
      \node (r0) at (2,-4.3){};
      \node (r1) at (2,-5) {};
      \draw[fill=red!10] 
      (12,-4.25)  -- (16,-4.25) -- (16,-3) -- (12,-3)--(12,-4.25); 
      \node[red] (r) at (14,-3.625) {OOB error};
      \node (r10) at (14,-4.3){};
      \node (r11) at (14,-5) {};
      \draw[fill=white] 
      (1,-6.2)  -- (3,-6.2) -- (3,-5) -- (1,-5)--(1,-6.2); 
      \node (r) at (2,-5.6) {$\err_1$};
      \draw[fill=white] 
      (5,-6.2)  -- (7,-6.2) -- (7,-5) -- (5,-5)--(5,-6.2); 
      \node (r) at (6,-5.6) {$\err_2$};
      \draw[fill=white] 
      (13,-6.2)  -- (15,-6.2) -- (15,-5) -- (13,-5)--(13,-6.2); 
      \node (r) at (14,-5.6) {$\err_B$};
      \node (r) at (8,-6.6) {$\underbrace{\qquad\qquad\qquad\qquad\qquad\qquad\qquad\qquad\qquad\qquad\qquad\qquad\qquad\qquad\qquad}$};
      \node (r) at (8,-8.0)
{$\err_{\oob}
=\cfrac{\err_1+\cdots+\err_B}{B}
=\cfrac{1}{B}{\displaystyle\,\sum_{b=1}^{B}\err_b}$};
      \draw[->]
      (t1) edge (t2) (t11) edge (t12) (s0) edge (s1) (s10) edge (s11) (r0) edge (r1) (r10) edge (r11);
  \end{tikzpicture}
  \vskip10pt
  \caption{\em Calculating the OOB prediction error for
    a model.  Blue points depict inbag sampled values,
    red points depict OOB values.  Model is fit using inbag data and
    then tested on OOB test data.  Averaging the prediction error over
  the different bootstrap realizations yields the OOB prediction error.}
  \label{F:err.oob}
\end{figure}

Calculating the OOB error begins with bootstrap sampling.  A bootstrap
sample is a sample of the data obtained by sampling with replacement.
Sampling by replacement creates replicated values.  On average, a
bootstrap sample contains only 63.2\% of the original data referred to
as in-sample or inbag.  The remaining 37\% of the data, which is
out-of-sample, and called the OOB data, is used as test data in the
OOB calculation.  The OOB error for a model is obtained by fitting a
model to bootstrap data, calculating its test set error on the OOB
data, and then repeating this many times ($B$ times), and averaging.
More technically, if $\err_b$ is the OOB test set error from the $b$th
bootstrap sample, the OOB error is
$$
\err_\oob = \frac{1}{B}\sum_{b=1}^B\err_b.
$$ 
See~\autoref{F:err.oob} for an illustration of calculating OOB error.

\setlength{\FrameRule}{2pt}
\setlength{\intextsep}{0pt}%
\setlength{\columnsep}{10pt}%

\subsection{Calculating the VIMP index for a variable}

The VIMP index for a variable $v$ is obtained by a slight extension to
the above procedure.  When calculating the OOB error for a model, the
OOB data for variable $v$ is ``noised up''.  Noising the OOB data is
intended to destroy the association between $v$ and the outcome.
Using the altered OOB data, one calculates the prediction error for
the model, call this $\err_{v,b}$ ($b$ is the specific bootstrap
sample). The VIMP index $\vimp_{v,b}$ is the difference between this
and the prediction error for the original OOB data, $\err_{b}$.  This
value will be positive if $v$ is predictive because the prediction
error for the noised up data will increase relative to the original
prediction error.  Averaging $\vimp_{v,b}$ yields the VIMP index,
$\vimp_{v}$,
$$
\vimp_v 
= \frac{1}{B}\sum_{b=1}^B \vimp_{v,b}  
= \frac{1}{B}\sum_{b=1}^B \[ \err_{v,b} - \err_b\].
$$
It follows that a positive value indicates a variable $v$ with a
predictive effect.  We call this value the {\it predictive effect
  size}.  A formal description of the VIMP algorithm is provided in
 Algorithm~\autoref{A:OOB.VIMP}.

\vskip15pt
\begin{algorithm}[htpb]
\centering
\caption{\em\,\, VIMP index for a variable $v$ \label{A:OOB.VIMP}}
\begin{algorithmic}[1]
  \FOR {$b=1,\ldots,B$}
  \STATE Draw a bootstrap sample of the data.
  \STATE Fit the model to the bootstrap data.
  \STATE Calculate the prediction error, $\err_b$, using the OOB data.
  \STATE Noise up the OOB data for $v$.
  \STATE Calculate the prediction error, $\err_{v,b}$, using the noised
    up OOB data.
  \STATE Calculate the boostrap VIMP index $\D_{v,b}=\err_{v,b} - \err_b$
  \ENDFOR
  \STATE Calculate the VIMP index by averaging: 
         $\vimp_v = \sum_{b=1}^B \vimp_{v,b} / B$. 
  \STATE The OOB error for the model can also be obtained using
         $\err_\oob = \sum_{b=1}^B\err_b/B$.
\end{algorithmic}
\end{algorithm}
\vskip10pt

We make several remarks regarding the implementation
of Algorithm~\autoref{A:OOB.VIMP}.
\Enumerate
\setlength\itemsep{-2pt}
\item
As stated, the algorithm provides a VIMP index for a given variable
$v$, but in practice one applies the same procedure for all variables
in the model.  The same bootstrap samples are to be used when doing
so.  This is required because it ensures that the VIMP index for each
variable is always compared to the same value $\err_b$.
\item
Because all calculations are run independently of one another, 
Algorithm~\autoref{A:OOB.VIMP} can be implemented using
parallel processing.  This makes the algorithm extremely fast and
scalable to big data settings.  The most obvious way to parallelize
the algorithm is on the bootstrap sample.  Thus, on a specific computing
machine on a cluster, a single bootstrap sample is drawn and $\err_b$
determined.  Steps 3-7 are then run for each variable in the model for
the given bootstrap draw.  Results from different computing machines
on the computing cluster are then averaged as in Steps 9 and 10.
\item
Noising up a variable is typically done by
permuting its data.  This approach is what is generaly used by
nonparametric regression models.  In the case of parametric and
semiparametric regression models (such as Cox regression), in place of
permutation noising up, the OOB data for the variable $v$ is set to
zero.  This is equivalent to setting the regression coefficient
estimate for $v$ to zero which is the convenient way of implementing
this procedure.   
\item
As a side effect, the algorithm can also be used to return the OOB
error rate for the model, $\err_{\oob}$ (see Step 10).  This can be useful for
assessing the effectiveness of the model and identifying 
poorly constructed models.  
\item
Algorithm~\autoref{A:OOB.VIMP} requires being able to calculate
prediction error.  The type of prediction error used will be context
specific.  For example in linear regression, prediction error can be
measured using mean-squared-error, or standardized mean-squared
errror.  In classification problems, prediction error is typically
defined by misclassification.  In survival problems, a common measure
of prediction performance is the Harrell's concordance index.  Thus
unlike the p-value, the interpretation of the VIMP index will be
context specific.  \EndEnumerate

\section{Risk factors for systolic heart failure}

To illustrate VIMP, we consider a survival data set
previously analyzed in~\cite{peak.v02}.  The data involves 2231
patients with systolic heart failure who underwent cardiopulmonary
stress testing at the Cleveland Clinic.  Of these 2231 patients,
during a mean follow-up of 5 years, 742 died.  In total, 39 variables
were measured for each patient including baseline characteristics and
exercise stress test results.  Specific details regarding the cohort,
exclusion criteria, and methods for collecting stress test data are
discussed in~\cite{peak.v02}.

We used Cox regression to fit the data using all cause mortality for
the survival endpoint (as was used in the original analysis). Only
linear variables were included in the model (i.e.\ no attempt was made
to fit non-linear effects).  Prediction error was assessed by the
Harrell's concordance index as described in~\cite{rsf}.  For improved
interpretation, prediction error was multiplied by 100.  This is
helpful because the resulting VIMP becomes expressible in terms of a
percentage.  For example, a VIMP index of 5\% indicates a variable
that improves by 5\% the ability of the model to rank patients by
their risk.  We should emphasize once again that VIMP is
cross-validated and provides a measure of predictive effect size.

\begin{table}[phtb]
\caption{\em Results from analysis of systolic heart failure data.}
\label{T:peakV02.vimp.index}
\vskip-10pt
\centering
\begin{tabular}{lrr|rrr|c}
\noalign{\hrule height 3.5pt}\\[-12pt]
\multicolumn{1}{l}{}&
\multicolumn{2}{c|}{Cox Regression}&
\multicolumn{3}{c|}{VIMP}&
\multicolumn{1}{c}{Marginal}\\
&&&&&&VIMP\\[6pt]
\multicolumn{1}{l}{Variable}&
\multicolumn{1}{c}{$\bhat$}&
\multicolumn{1}{c|}{p-value}&
\multicolumn{1}{c}{$\bbag$}&
\multicolumn{1}{c}{$\vimp_\b$}&
\multicolumn{1}{c|}{$\err_\step$}&
\multicolumn{1}{c}{$\vimp_\b^\marg$}\\[2pt]
\noalign{\hrule height 1.5pt}\\[-8pt]
Peak VO$_2$ & -0.06 & 0.002 & -0.06 & 1.94 & 32.40 & 0.25 \\ 
  BUN & 0.02 & 0.000 & 0.02 & 1.67 & 30.81 & 0.37 \\ 
  Exercise time & 0.00 & 0.008 & 0.00 & 1.37 & 30.80 & 0.08 \\ 
  Male & 0.47 & 0.000 & 0.47 & 0.52 & 30.01 & 0.37 \\ 
  beta-blocker & -0.23 & 0.006 & -0.23 & 0.30 & 29.34 & 0.16 \\ 
  Digoxin & 0.36 & 0.000 & 0.36 & 0.30 & 29.00 & 0.22 \\ 
  Serum sodium & -0.02 & 0.071 & -0.02 & 0.20 & 28.93 & 0.07 \\ 
  Age & 0.01 & 0.022 & 0.01 & 0.18 & 28.99 & -0.03 \\ 
  Resting heart rate & 0.01 & 0.058 & 0.01 & 0.14 & 28.93 & 0.04 \\ 
  Angiotensin receptor blocker & 0.26 & 0.067 & 0.27 & 0.13 & 28.92 & 0.02 \\ 
  LVEF & -0.01 & 0.079 & -0.01 & 0.11 & 28.86 & 0.03 \\ 
  Aspirin & -0.21 & 0.018 & -0.21 & 0.11 & 28.83 & 0.03 \\ 
  Resting systolic blood pressure & 0.00 & 0.158 & 0.00 & 0.07 & 28.83 & 0.00 \\ 
  Diabetes insulin treated & 0.26 & 0.057 & 0.25 & 0.07 & 28.87 & -0.02 \\ 
  Previous CABG & 0.11 & 0.316 & 0.12 & 0.07 & 28.86 & -0.02 \\ 
  Coronary artery disease & 0.12 & 0.284 & 0.12 & 0.06 & 28.92 & -0.04 \\ 
  Body mass index & 0.00 & 0.800 & 0.00 & 0.00 & 28.96 & -0.05 \\ 
  Potassium-sparing diuretics & -0.14 & 0.134 & -0.14 & -0.03 & 28.97 & -0.01 \\ 
  Previous MI & 0.29 & 0.012 & 0.30 & -0.03 & 29.02 & -0.01 \\ 
  Thiazide diuretics & 0.04 & 0.707 & 0.04 & -0.04 & 29.07 & -0.05 \\ 
  Peak respiratory exchange ratio & 0.12 & 0.701 & 0.12 & -0.04 & 29.12 & -0.05 \\ 
  Statin & -0.12 & 0.183 & -0.13 & -0.04 & 29.19 & -0.07 \\ 
  Antiarrythmic & 0.04 & 0.700 & 0.04 & -0.04 & 29.25 & -0.06 \\ 
  Diabetes noninsulin treated & 0.01 & 0.930 & 0.00 & -0.05 & 29.30 & -0.06 \\ 
  Dihydropyridine & 0.03 & 0.851 & 0.03 & -0.05 & 29.35 & -0.05 \\ 
  Serum glucose & 0.00 & 0.486 & 0.00 & -0.05 & 29.42 & -0.07 \\ 
  Previous PCI & -0.06 & 0.557 & -0.06 & -0.05 & 29.48 & -0.05 \\ 
  ICD & 0.04 & 0.676 & 0.03 & -0.05 & 29.55 & -0.07 \\ 
  Anticoagulation & -0.01 & 0.933 & -0.01 & -0.06 & 29.61 & -0.06 \\ 
  Pacemaker & -0.02 & 0.851 & -0.01 & -0.06 & 29.67 & -0.06 \\ 
  Current smoker & 0.03 & 0.807 & 0.03 & -0.06 & 29.74 & -0.06 \\ 
  Nitrates & -0.04 & 0.623 & -0.04 & -0.06 & 29.80 & -0.06 \\ 
  Serum hemoglobin & 0.00 & 0.923 & 0.01 & -0.06 & 29.87 & -0.07 \\ 
  Black & 0.07 & 0.589 & 0.06 & -0.07 & 29.95 & -0.08 \\ 
  Nondihydropyridine & -0.30 & 0.510 & -0.51 & -0.07 & 30.03 & -0.08 \\ 
  Loop diuretics & -0.07 & 0.541 & -0.08 & -0.07 & 30.09 & -0.06 \\ 
  ACE inhibitor & 0.10 & 0.371 & 0.11 & -0.09 & 30.15 & -0.06 \\ 
  Vasodilators & -0.08 & 0.606 & -0.07 & -0.09 & 30.25 & -0.09 \\ 
  Creatinine clearance & 0.00 & 0.624 & 0.00 & -0.11 & 30.31 & -0.06 \\ 
\\[-8pt]\noalign{\hrule height 1.5pt}
\end{tabular}
\end{table}

\autoref{T:peakV02.vimp.index} presents the results from the Cox
regression analysis.  Included are VIMP indices and other quantities
obtained from $B=1000$ bootstrapped Cox regression models.  Column
$\bhat$ lists the coefficient estimate for each $\b$ variable, $\bbag$
is the averaged coefficient estimate from the $1000$ bootstrapped
models.  Column 4 agrees closely with column 2, which is to be
expected if the number of iterations $B$ is selected suitably large.
\autoref{T:peakV02.vimp.index} has been sorted in terms of the VIMP
index, $\vimp_\b$.  Interestingly, ordering by VIMP does not match
ordering by p-value.  For example, insulin treated diabetes has a near
significant p-value of 6\%, however, its VIMP of 0.07\% is relatively
small compared with other variables.  The top variable peak VO$_2$ has
a VIMP of 1.9\%, which is over 27 times larger.

Peak VO$_2$, BUN, and treadmill exercise time are the top three
variables identified by the VIMP index.  Following these
are an assortment of variables with moderate VIMP: sex, use of
beta-blockers, use of digoxin, serum sodium level, and age of patient.
Then are variables with small but non-zero VIMP, starting with patient
resting heart rate, and terminating with presence of coronary artery
disease.  VIMP indices become zero or negative after this.  These
latter variables, with zero or negative VIMP indices, can be viewed as
``noisy'' variables that degrade model performance.  This can be seen
by considering the column labeled as $\err_\step$.  This equals the
OOB prediction error for each stepwise model ordered by VIMP.  
\autoref{T:stepwise} lists the 
stepwise models that were considered. For
example the third line, 30.80, is the OOB error for the model using
top three variables.  The fourth line is the OOB error for the top
four variables, and so forth.  \autoref{T:peakV02.vimp.index} shows
that $\err_\step$ decreases for models with positive VIMP, but rises
once models begin to include noisy variables with zero or negative
VIMP.

\Remark
Because prediction error will be optimistic for models based on ranked
variables, we calculate $\err_\step$ using the same bootstrap samples
used by Algorithm~\autoref{A:OOB.VIMP}.  Thus, the value 30.31 in the
last row of column $\err_\step$, corresponding to fitting the entire
model, coincides exactly with the OOB model prediction error obtained
using Algorithm~\autoref{A:OOB.VIMP}.  
\EndRemark

\vskip8pt
\begin{table}[phtb]
\begin{framed}
\caption{\em Stepwise models used in calculating $\err_\step$.}
\label{T:stepwise}
\vskip-10pt
$$
\begin{array}{L@{\quad}L@{\quad}}
\underline{Model Number} & \underline{Stepwise Model}\\[6pt]
1 & Model using the top variable only, $\{\text{peak VO$_2$}\}$\\
2 & Model using top two variables, $\{\text{peak VO$_2$, BUN}\}$\\
3 & Model using top three variables, $\{\text{peak VO$_2$, BUN,
exercise time}\}$\\
\vdots    & \vdots\\
39& Model using all 39 variables
\end{array}
$$ 
\end{framed}
\end{table}

\section{Marginal VIMP}

Now we explain the meaning of the column entry $\vimp_\b^\marg$
in~\autoref{T:peakV02.vimp.index}.  Recall that $\err_\step$ measures
the OOB prediction error for a specific stepwise model.  Relative to
its previous entry, it estimates the effect of a variable when added
to the current model.  For example, the effect of adding exercise time
to the model with peak VO$_2$ and BUN is the difference between the
second row (model 2), 30.81, and the third row (model 3), 30.80.  The
effect of adding exercise time is therefore 0.01 (30.81 minus 30.80).
This is much smaller than the VIMP index for exercise time which
equals 1.37.  These values differ because the stepwise error rate
estimates the effect of {\it adding} treadmill exercise time to the
model with Peak VO$_2$ and BUN.  We call this the {\it discovery
  effect size} of the variable.  The entry $\vimp_\b^\marg$
in~\autoref{T:peakV02.vimp.index} is a generalization of this concept
and is what we call the marginal VIMP.  It calculates the discovery
effect of a variable compared to the model containing all variables
{\it except} that variable.  
\autoref{marginal.versus.vimp} summarizes the difference
between marginal VIMP and the VIMP index.

\vskip15pt
\begin{table}[phtb]
\begin{framed}
\caption{\em Difference between VIMP and marginal VIMP.}
\label{marginal.versus.vimp}
\centering\vskip5pt 
VIMP is calculated through noising up a
variable.\\ 
Marginal VIMP is calculated through removing a variable.
\vskip10pt
\caption*{\em Note that removing a variable from the model will
  change the coefficients of other variables, while noising up a
  variable will not change those.}
\vskip-10pt
\end{framed}
\end{table}

The marginal VIMP is easily calculated by a simple modification to
Algorithm~\autoref{A:OOB.VIMP}.  In place of noising up a variable
$v$, a second model is fit to the bootstrap data, but with $v$
removed.  The OOB error for this model is compared
to the OOB error for the full model containing all variables.
Averaging these values over the bootstrap realizations yields
$\vimp_v^\marg$. See Algorithm~\autoref{A:OOB.marginal.VIMP} for a
formal description of this procedure.
\vskip20pt

\vskip5pt
\begin{algorithm}[htpb]
\centering
\caption{\em\,\, Marginal VIMP for a variable $v$ \label{A:OOB.marginal.VIMP}}
\begin{algorithmic}[1]
  \FOR {$b=1,\ldots,B$}
  \STATE Draw a bootstrap sample of the data.
  \STATE Fit the model to the bootstrap data and calculate its
  prediction error, $\err_b$, using the OOB data.
  \STATE Fit a second model, but without variable $v$, and calculate
  its predictiction error, $\err_{v,b}^\marg$ using the OOB data.
  \ENDFOR
  \STATE Calculate the marginal VIMP by averaging: 
         $\vimp_v^\marg = \sum_{b=1}^B \[\err_{v,b}^\marg-\err_b\] / B$. 
\end{algorithmic}
\end{algorithm}
\vskip15pt

\begin{figure}[pht]
\bct
\resizebox{2.50in}{!}{\includegraphics[page=1]{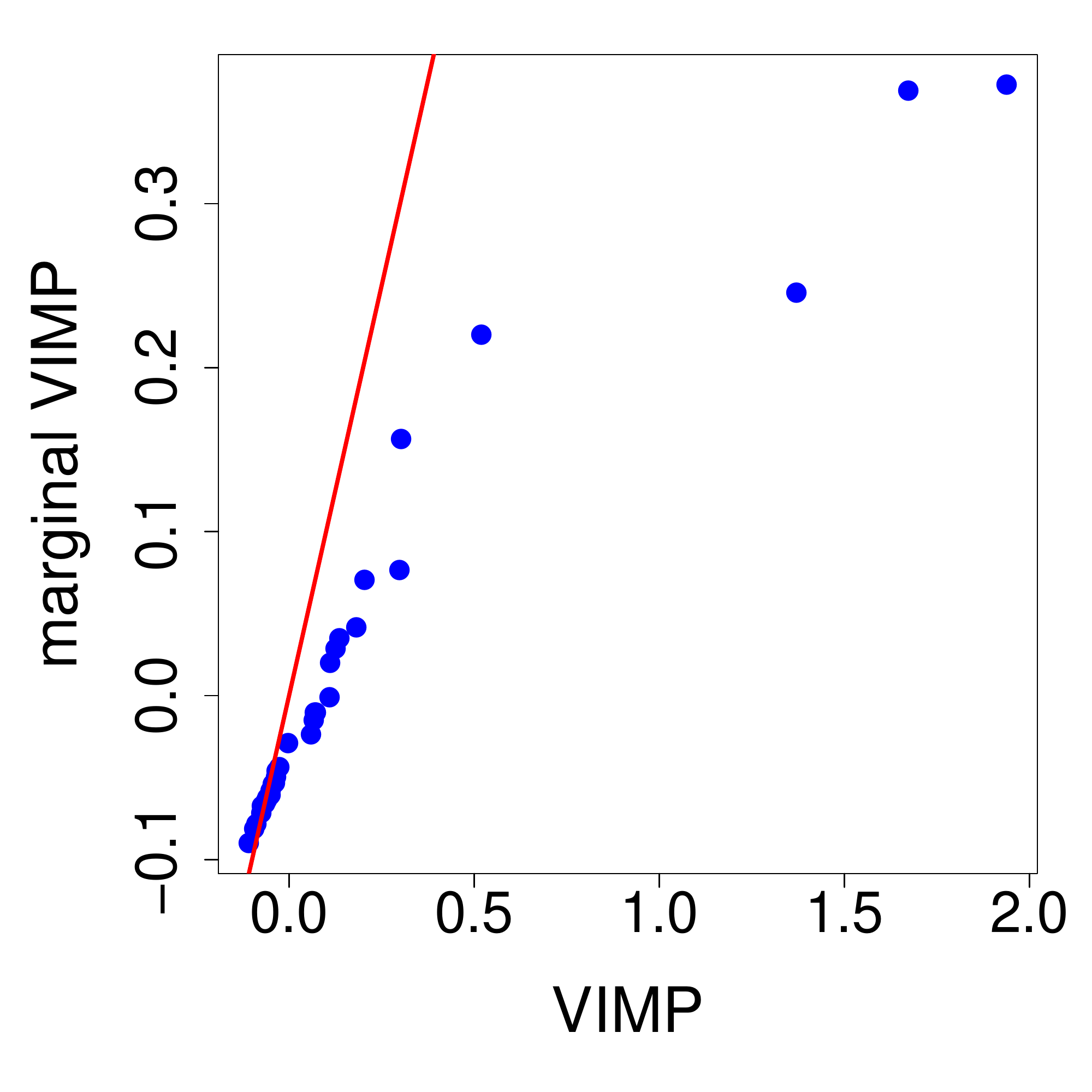}}
\resizebox{2.50in}{!}{\includegraphics[page=2]{peakV02_compare.pdf}}
\ect
\vskip-10pt
\caption{\em Differences between marginal VIMP and the VIMP index for
  systolic heart failure data.
  Left-hand figure displays the two values plotted against each other.
  Right-hand figure compares the ranking of variables by the two methods.}
\label{F:peakV02.cox}
\end{figure}

\autoref{T:peakV02.vimp.index} reveals interesting differences between
marginal VIMP and the VIMP index.  Generally, marginal VIMP is much
smaller.  We can conclude that the discovery effect size is a
conservative measure, as we would expect given the large number of
variables in our model.  Second, as expected, the discovery effect of
exercise time is substantially smaller than its VIMP index.  Third,
there is a small collection of variables whose discovery effect is
relatively large compared to their VIMP index.  The most interesting
is sex, which has the largest discovery effect among all variables
(being tied with BUN).  The explanation for this is that adding sex to
the model supplies ``orthogonal'' information not contained in other
variables.  Marginal VIMP is in some sense a statement about
correlation.  For example, correlation of exercise time with peak
VO$_2$ is 0.87, whereas correlation of BUN with peak VO$_2$ is -0.40.
This allows BUN to have a high discovery effect when peak VO$_2$ is
included in the model, while exercise time cannot.  Differences
between marginal VIMP and VIMP indices are summarized
in~\autoref{F:peakV02.cox}.  The right-hand plot displays the ranking
of variables by the two methods.  There is some overlap in the top
variables (points in lower left hand side), but generally we see
important differences.

\section{Robustness of VIMP to the sample size}

\begin{figure}[pht]
\bct
\resizebox{6.0in}{!}{\includegraphics[page=2]{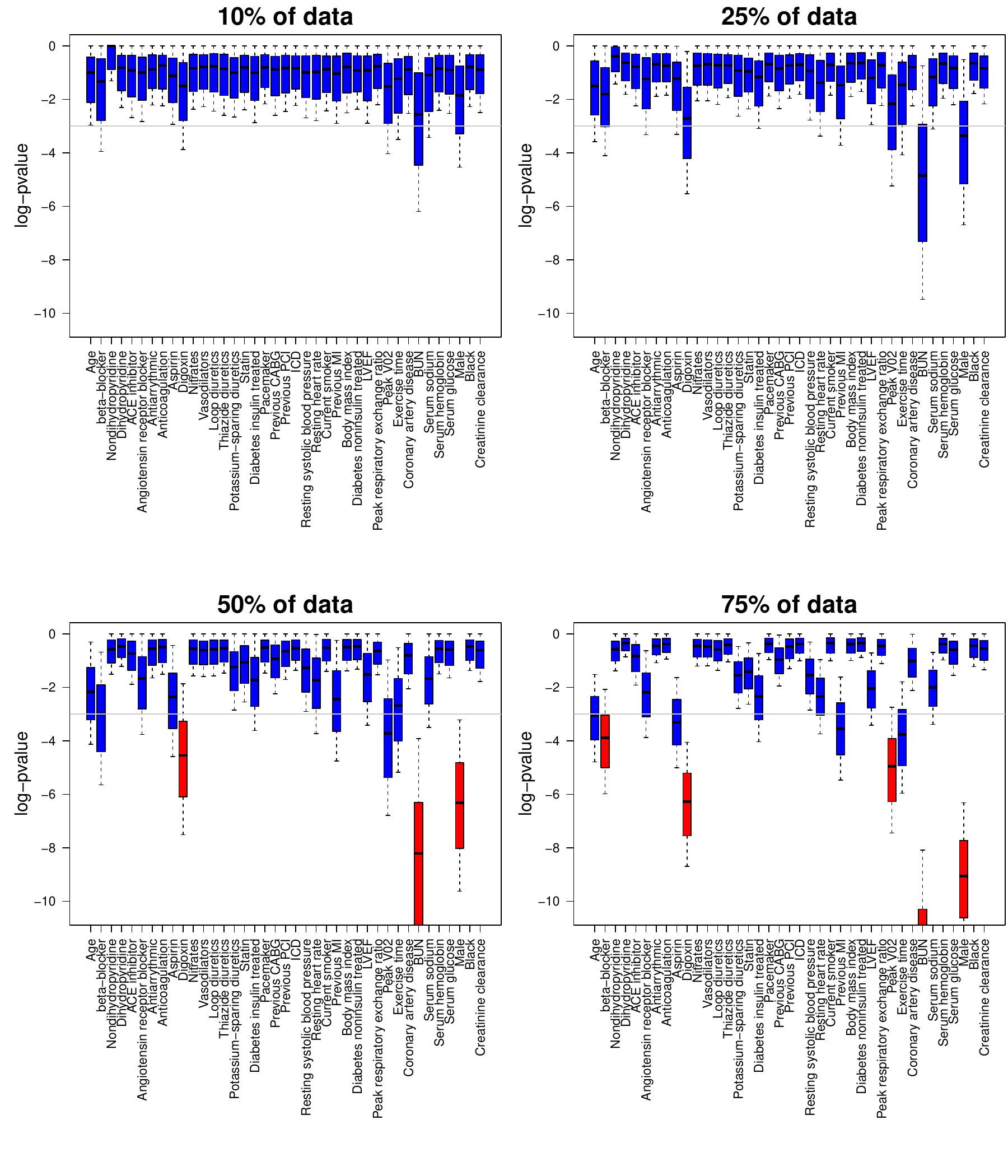}}
\ect
\vskip-10pt
\caption{\em Logarithm of p-value as a function of fraction of sample
  size for systolic heart failure data (large negative values
  correspond to near zero p-values).  Values are calculated using 500
  independently subsampled data sets.  Horizontal line is
  $\log(0.05)$, the typical threshold used to identify a significant
  variable.}
\label{F:pvalue.sample.size}
\end{figure}

\begin{figure}[pht]
\bct
\resizebox{6.0in}{!}{\includegraphics[page=3]{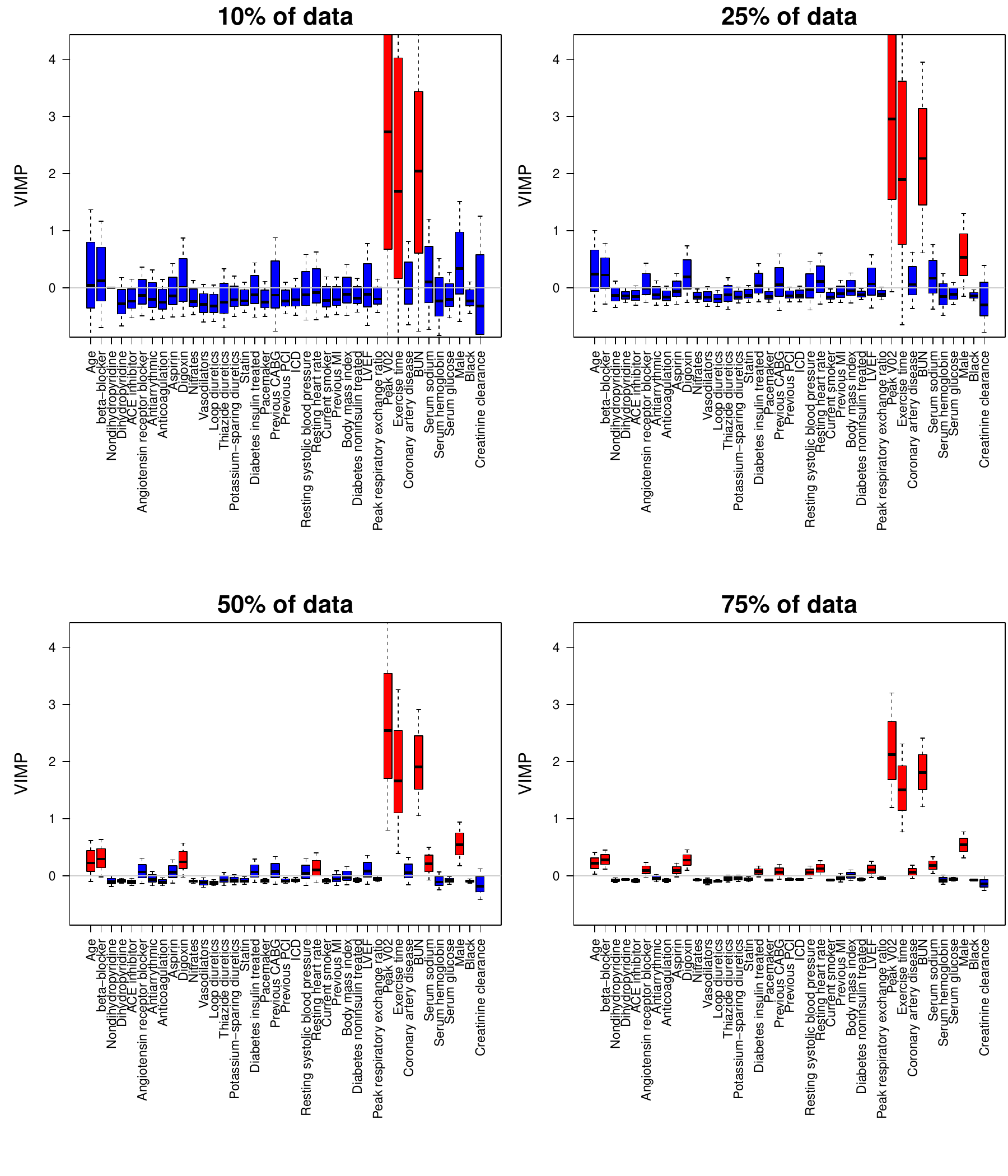}}
\ect
\vskip-10pt
\caption{\em Subsampled data is the same as Figure~\ref{F:pvalue.sample.size} but
  where VIMP is now reported.}
\label{F:vimp.sample.size}
\end{figure}

Here we demonstrate the robustness of VIMP to the sample size.  We use
the systolic heart failure data as before, but this time using only a
fraction of the data.  We used a random 10\%, 25\%, 50\%, and 75\% of
the data.  This process was repeated 500 times independently.  For
each data set, we saved the p-values and VIMP indices for all
variables.  \autoref{F:pvalue.sample.size} displays the logarithm of
the p-values from the experiment (large negative values correspond to
near zero p-values). \autoref{F:vimp.sample.size} displays the VIMP
indices.  What is most noticeable from \autoref{F:vimp.sample.size} is
that VIMP indices are informative even in the extremely low sample
size setting of 10\%.  For example, VIMP interquartile values (the
lower and upper ends of the boxplot) are above zero for peak VO$_2$,
BUN, and treadmill exercise time, showing that VIMP is able to
consistently identify the top three variables even with limited data.
In contrast, in \autoref{F:pvalue.sample.size} for the low sample
setting of 10\%, no variable had a median log p-value below the
threshold of $\log(0.05)$; showing that no variable met the 5\% level
of significance on average.  Furthermore, even with 75\% of the data,
the upper end of the boxplot for exercise time is still above the
threshold, showing its significance is questionable.  These results
demonstrate the robustness of VIMP to sample size in contrast to the
p-value.

\section{Misspecified model}

\begin{figure}[pht]
\bct
\resizebox{2.75in}{!}{\includegraphics[page=1]{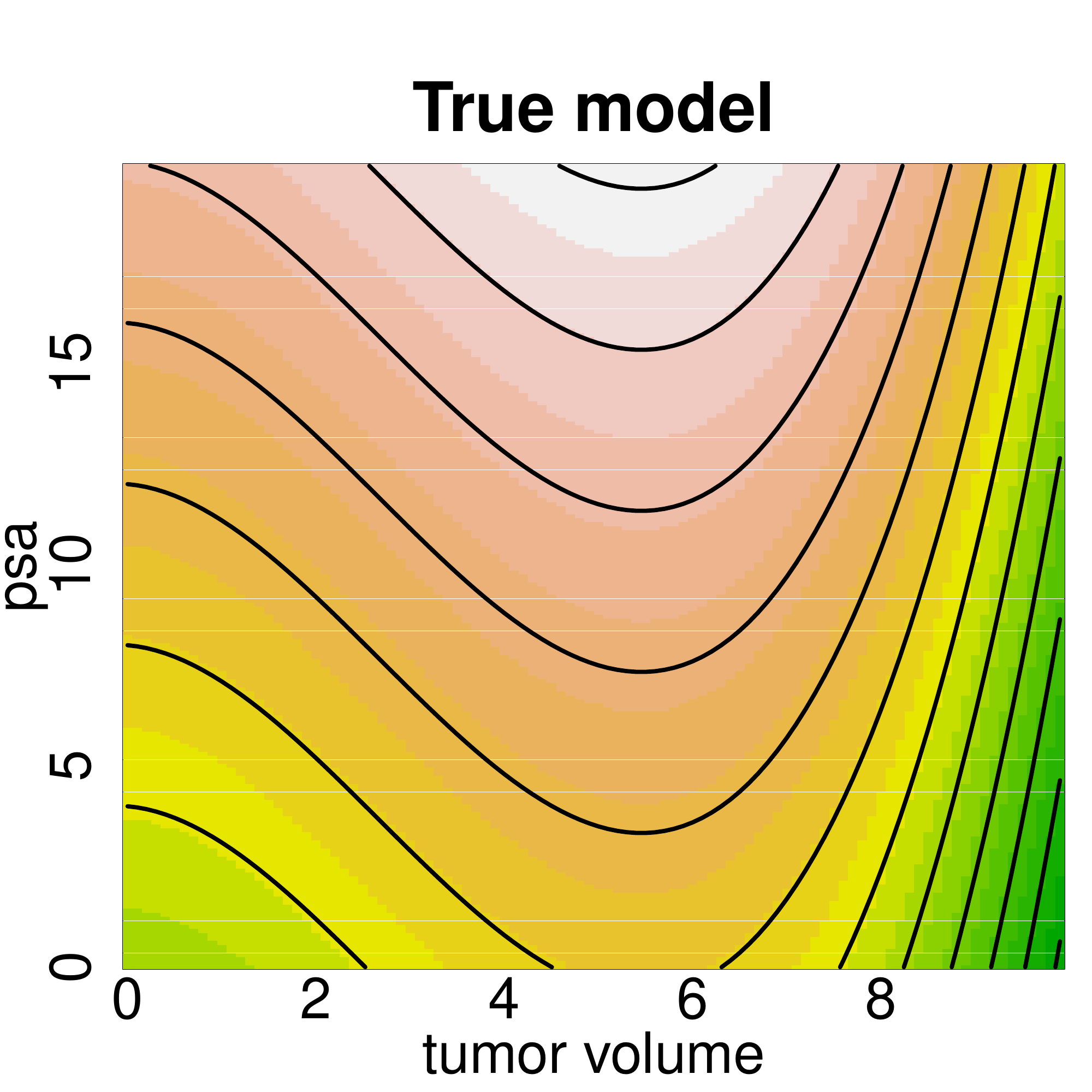}}
\resizebox{2.75in}{!}{\includegraphics[page=2]{survival_simulation_contour.pdf}}
\ect
\vskip-10pt
\caption{\em Log-hazard function from Cox simulation example.  Left figure
  displays the true log-hazard function which includes the non-linear term
  for tumor volume. Right figure displays the log-hazard function assuming
  linear variables only.}
\label{F:survival.simulation}
\end{figure}

For our next illustration we used simulations to demonstrate
robustness of VIMP to model misspecification.  For our simulation, we
sampled $n=1000$ values from a Cox regression model with five
variables.  The first two variables are ``psa'' and ``tumor volume''
and represent variables associated with the survival outcome.  The
remaining three variables are noise variables with no relationship to
the outcome.  These are called $X_1,X_2,X_3$.  The variable psa has a
linear main effect, but tumor volume has both a linear and non-linear
term.  The true regression coefficient for psa is 0.05 and the
coefficient for the linear term in tumor volume is 0.01.  A censoring
rate of approximately 70\% was used.  The log of the hazard function
used in our simulation is given in the left panel
of~\autoref{F:survival.simulation}.  Mathematically, our log-hazard
function assumes the following function
$$
\log(h(t)) = \a_0 + 0.05 \times \text{psa} + 0.01 \times \text{tumor volume}
 + \psi(\text{tumor volume})
$$ 
where $\psi(x)=0.04x^2-0.005x^3$ is a polyomial function with
quadratic and cubic terms.  The right panel
of~\autoref{F:survival.simulation} displays the log-hazard for the
misspecified model that does not include the non-linear term for
tumor volume.

\vskip15pt
\vskip15pt

We first fit a Cox regression model to the data using only linear
variables as one might typically do. Following this,
Algorithms~\autoref{A:OOB.VIMP} and~\autoref{A:OOB.marginal.VIMP} were
applied with $B=1000$. The entire procedure was then repeated $M=1000$
times.  Each of these Monte Carlo runs consisted of simulating a new
data set, fitting a Cox regression model to this simulated data, and
running Algorithms~\autoref{A:OOB.VIMP}
and~\autoref{A:OOB.marginal.VIMP}.  The results are
summarized in~\autoref{T:survival.vimp}.  All reported values are
averaged over the $M=1000$ Monte Carlo experiments.

We first fit a Cox regression model to the data using only linear
variables as one might typically do.  This model was bootstrapped
$B=1000$ values and VIMP and marginal VIMP calculated.  This entire
procedure of simulating a data set, fitting a Cox model and 1000
bootstrapped Cox models, was repeated $M=1000$ times.  The results
from these 1000 Monte Carlo experiments were averaged.  These values
are summarized in~\autoref{T:survival.vimp}.  The table shows that the
p-value has no difficulty in identifying the strong effect of psa,
which is correctly specified in the model.  However, the p-value for
tumor volume is 0.267, indicating a non-significant effect.  The
p-value tests whether this coefficient is zero, assuming the model is
true, but the problem is that the fitted model is misspecified.  The
estimated Cox regression model inflates the coefficient for tumor
volume in a negative direction (estimated value of -0.03, but true
value is 0.01) in an attempt to compensate for the non-linear effect
that was excluded from the model.  This leads to the invalid p-value.
In contrast, both the VIMP and marginal VIMP values for tumor volume
are positive.  Although these values are substantially smaller than
the values for psa, VIMP is still able to identify a predictive
effect size associated with tumor volume.  Once again, this is
possible because VIMP bases its estimation on test data and not a
presumed model which can be incorrect.  Also, notice that all three
noise variables are correctly identified as uninformative.  All have
negative VIMP values.

\vskip15pt
\begin{table}[phtb]
\caption{\em Results from analysis of simulated Cox regression data
  set.  The model is misspecified by failing to include the
non-linear term for tumor volume.}
\label{T:survival.vimp}
\centering
\begin{tabular}{lrr|rrc}
\noalign{\hrule height 3.5pt}\\[-12pt]
\multicolumn{1}{l}{}&
\multicolumn{1}{c}{$\bhat$}&
\multicolumn{1}{c|}{p-value}&
\multicolumn{1}{c}{$\bbag$}&
\multicolumn{1}{c}{$\vimp_\b$}&
\multicolumn{1}{c}{$\vimp_\b^\marg$}\\[2pt]
\noalign{\hrule height 1.5pt}\\[-8pt]
  psa & 0.05 & 0.001 & 0.05 & 6.32 & 6.34 \\ 
tumor volume & -0.03 & 0.267 & -0.03 & 0.14 & 0.15 \\ 
  $X_1$ & 0.00 & 0.490 & 0.00 & -0.25 & -0.25 \\ 
  $X_2$ & 0.00 & 0.486 & 0.00 & -0.25 & -0.25 \\ 
  $X_3$ & 0.00 & 0.493 & 0.00 & -0.27 & -0.27 \\ 
\\[-8pt]
\noalign{\hrule height 1.5pt}
\end{tabular}
\\\vskip5pt
{\footnotesize The overall OOB model error is 43\%.}
\end{table}
\vskip20pt

Typically, a standard analysis would end after looking at the
p-values.  However, a researcher with access to the entire
\autoref{T:survival.vimp} might be suspicious of the small positive
VIMP of tumor volume and its negative coefficient value which is
unexpected from previous experience.  This combined with the high OOB
model error (equal to 43\%) should alert them to consider more
sophisticated modeling.  This is easily done using standard
statistical methods.  Here we use B-splines~\citep{b-splines} to add
non-linearity to tumor volume.  This expands the design matrix for the
Cox regression model to include additional columns for the B-spline
expansion of tumor volume.  When noising up tumor volume all of these
B-spline columns are noised up simultaneously (i.e.\ their coefficient
estimates are set to zero).
The extensions to
Algorithms~\autoref{A:OOB.VIMP} and~\autoref{A:OOB.marginal.VIMP} are
straightforward.

\vskip15pt
\begin{table}[ht]
\caption{\em Results from Cox regression simulation using
  a B-spline to model non-linearity in tumor volume.}
\label{T:survival.vimp.bs}
\centering
\begin{tabular}{lrr}
\noalign{\hrule height 3.5pt}\\[-12pt]
\multicolumn{1}{l}{}&
\multicolumn{1}{c}{$\vimp_\b$}&
\multicolumn{1}{c}{$\vimp_\b^\marg$}\\[2pt]
\noalign{\hrule height 1.5pt}\\[-8pt]
  psa & 4.20 & 4.23 \\ 
tumor volume & 2.27 & 2.31 \\ 
  $X_1$ & -0.20 & -0.20 \\ 
  $X_2$ & -0.20 & -0.20 \\ 
  $X_3$ & -0.21 & -0.21 \\ 
\\[-8pt]
\noalign{\hrule height 1.5pt}
\end{tabular}
\\ \vskip5pt
{\footnotesize The overall OOB model error is 40\%.}
\end{table}
\vskip20pt

The results from the B-spline analysis are displayed
in~\autoref{T:survival.vimp.bs}.  As before, the entire procedure was
repeated $M=1000$ times, with values averaged over the Monte Carlo
runs.  Notice the large values of VIMP for tumor volume.  The overall
model performance has also improved to 40\%.  Overall, results
have improved substantially.

\section{Discussion}

It seems questionable that the p-value can continue to meet the needs
of scientists.  It does not provide an interpretable scientific effect
size that researchers desire and it is valid only if the underlying
model holds, which can often be questionable given the restrictive
assumptions often used with traditional modeling.  In this paper, we
introduced VIMP as an alternative approach.  VIMP provides an
interpretable measure of effect size that is robust to model
misspecification.  It uses prediction error based on out-of-sample
data and replaces statistical significance with predictive importance.
The VIMP framework is feasible to all kinds of models including not
only parametric models, such as those considered here, but also
non-parametric models such as those used in machine learning
approaches.

We discussed two types of VIMP measures: the VIMP index and the
marginal VIMP.  The scientific application will dictate which of these
is more suitable.  VIMP indices are appropriate in settings where
variables for the model are already established and the goal is to
identify the predictive effect size. For example, if several genetic
markers are already identified as a genetic cause for coronary heart
disease risk, VIMP can provide a rank for these and estimate the
magnitude each marker plays in the prediction for the outcome.
Marginal VIMP is appropriate when the goal is new scientific
discovery. For instance, if a researcher is proposing to add a new
genetic marker for evaluating coronary heart disease risk, marginal
VIMP can yield a discovery effect size for how much the new proposed
marker adds to previous risk models.

From a statistical perspective, VIMP idices
are an OOB alternative to the regression coefficient p-value.
However, what VIMP measures about a variable can be very flexible.  It
may be a linear effect, or quite easily a non-linear effect, such as
modeled using B-splines.  An important feature is that degrees of
freedom and other messy details required with p-values when dealing
with complex modeling are never an issue with VIMP. Marginal VIMP is an OOB analog to the
likelihood-ratio test. In statistics, likelihood-ratio tests compare
the goodness-of-fit of two models, one of which (the null model with
certain variables removed) is a special case of the other (the
alternative model with all variables included).  Marginal VIMP
compares the prediction precision of these two scenarios. 

Because both VIMP and marginal VIMP are measures of predictive
importance, their values are standardized to the measure of prediction
performance used.  This makes it possible to compare values across
different data sets.  For example, a 0.05 VIMP value for two different
variables from two different survival datasets is comparable---both
imply a 5\% contribution to the concordance index.  Another feature
which we touched upon briefly in our B-spline example is the ability
to use VIMP to measure the effect of groups of variables.  In our
B-spline example, the cluster of variables used were the B-spline
contributions to tumor volume, and were combined together to give an
overall estimate of the effect of tumor volume.  One could easily
extend this to calculate cluster-VIMP as a better sense of the
importance of a highly correlated group of variables.


\begin{thebibliography}{}

\bibitem[Breiman et al.(1984)]{CART}
Breiman L., Friedman J.H., Olshen R.A., and Stone C.J.
\newblock {\em Classification and Regression Trees}.
\newblock Wadsworth, Belmont, California, 1984.

\bibitem[Breiman(1996)]{oob}
Breiman, L.
\newblock Out-of-bag estimation. 
Technical report, Statistics Dept., University of California at
Berkeley, CA, 1996.

\bibitem[Breiman(1996)]{bagging}
Breiman, L.
\newblock Bagging predictors.
\newblock {\em Machine learning}, 24(2), 123--140, 1996.

\bibitem[Breiman(2001a)]{two-cultures}
Breiman, L.
\newblock Statistical modeling: The two cultures (with comments and a
rejoinder by the author). 
\newblock {\em Statistical Science}, 16(3), pp.199--231, 2001a.

\bibitem[Breiman(2001b)]{random.forest}
Breiman, L.
\newblock Random forests.
\newblock {\em Machine learning}, 45(1), 5--32, 2001b.

\bibitem[Efron(1983)]{efron:error-rate}
Efron B.
\newblock Estimating the error rate of a prediction rule: improvement on
  cross-validation.
\newblock {\em J. Amer. Stat. Assoc.}, 78:316--331, 1983.

\bibitem[Efron and Tibshirani(1997)]{632}
Efron B. and Tibshirani R.
\newblock Improvements on cross-validation: the .632+ bootstrap method.
\newblock {\em J. Amer. Stat. Assoc.}, 92:548--560, 1997.

\bibitem[Eilers and Marx(1996)]{b-splines}
Eilers~P.H.C. and Marx~B.D.
\newblock Flexible smoothing with {B-splines} and penalties.
\newblock {\em Statistical Science}, pages 89--102, 1996.

\bibitem[Hsich et al.(2011)]{peak.v02}
Hsich E., Gorodeski E.Z., Blackstone E.H., Ishwaran H., and Lauer M.S.
\newblock Identifying important risk factors for survival in systolic heart
  failure patients using random survival forests.
\newblock {\em Circ. Cardiovasc. Qual. Outcomes}, 4(1):39--45, 2011.

\bibitem[Ishwaran et al.(2008)]{rsf}
Ishwaran H., Kogalur U.B., Blackstone E.H., and Lauer M.S.
\newblock Random survival forests.
\newblock {\em Annals of Applied Statistics}, 2(3):841--860, 2008.

\bibitem[Nuzzo(2014)]{nature}
Nuzzo, R. (2014), 
\newblock Scientific method: statistical errors.
\newblock{Nature}, 506, 150--152, 2014.\\
\url{www.nature.com/news/scientific-method-statistical-errors-1.14700}

\bibitem[Trafimow and Marks(2015)]{ban-pvalue}
Trafimow, D. and Marks, M.
\newblock Editorial.
\newblock{\em Basic and Applied Social Psychology}, 37(1):1--2, 2015.

\bibitem[Wasserstein and Lazar(2016)]{asa}
Wasserstein, R.L. and Lazar, N.A. 
\newblock The ASA's statement on p-values: context, process, and purpose. 
\newblock {\em The American Statistician}, 70, 129--133, 2016.


\end{thebibliography}
\end{document}